\documentclass[conference]{IEEEtran}
\usepackage{times}

\usepackage[numbers]{natbib}
\usepackage{multicol}
\usepackage[bookmarks=true]{hyperref}

\usepackage{amsmath}
\usepackage{amsfonts}
\usepackage{amsthm}
\usepackage{graphicx}
\usepackage{subfig}
\pdfoutput=1

\DeclareMathOperator*{\argmax}{arg\,max}

\usepackage[textfont=md,font=footnotesize]{caption}
\renewcommand{\baselinestretch}{.99}
\newcommand{\eref}[1]{(\ref{#1})}
\newcommand{\sref}[1]{Sec. \ref{#1}}
\newcommand{\figref}[1]{Fig. \ref{#1}}

\newcommand{\prg}[1]{\noindent\textbf{#1. }} 

\newcommand{\tightoverset}[2]{%
  \overset{#1}{#2}}

\newcommand{\thetaprox}{{\ensuremath{\tightoverset{\sim}{\theta}}}}
\newcommand{\thetatrue}{\ensuremath{\theta^*}}
\newcommand{\trainenv}{{\ensuremath{\tightoverset{\sim}{M}}}}

\usepackage{color}

\newcommand{\adnote}[1]%
 {\textcolor{blue}{\textbf{AD: #1}}} 

\begin{document}
\title{Simplifying Reward Design \\through Divide-and-Conquer}

% \author{Author Names Omitted for Anonymous Review. Paper-ID 158}

\author{\authorblockN{Ellis Ratner \qquad Dylan Hadfield-Menell \qquad Anca D. Dragan}
\authorblockA{%Department of Electrical Engineering and Computer Science \\
%University of California, Berkeley \\
%Email: 
\{\href{eratner@berkeley.edu}{eratner}, \href{dhm@berkeley.edu}{dhm}, \href{anca@berkeley.edu}{anca}\}@berkeley.edu}}

%\renewcommand{\shortauthors}{E. Ratner et al.}

\maketitle

\begin{abstract}
Designing a good reward function is essential to robot planning and reinforcement learning, but it can also be challenging and frustrating. The reward needs to work across multiple different environments, and that often requires many iterations of tuning. We introduce a novel divide-and-conquer approach that enables the designer to specify a reward separately for each environment. By treating these separate reward functions as observations about the underlying true reward, we derive an approach to infer a common reward across all environments. We conduct user studies in an abstract grid world domain and in a motion planning domain for a 7-DOF manipulator that measure user effort and solution quality. We show that our method is faster, easier to use, and produces a higher quality solution than the typical method of designing a reward jointly across all environments. We additionally conduct a series of experiments that measure the sensitivity of these results to different properties of the reward design task, such as the number of environments, the number of feasible solutions per environment, and the fraction of the total features that vary within each environment. We find that independent reward design outperforms the standard, joint, reward design process but works best when the design problem can be divided into simpler subproblems.
\end{abstract}

%\IEEEpeerreviewmaketitle

% \blfootnote{*Confidential, under submission.}

\section{Introduction}

While significant advances have been made in planning and reinforcement learning for robots, these algorithms require access to a reward (or cost) function in order to be successful. Unfortunately, designing a good reward function by  hand remains challenging in many tasks. 

\begin{figure}[t!]
    \includegraphics[width=\columnwidth]{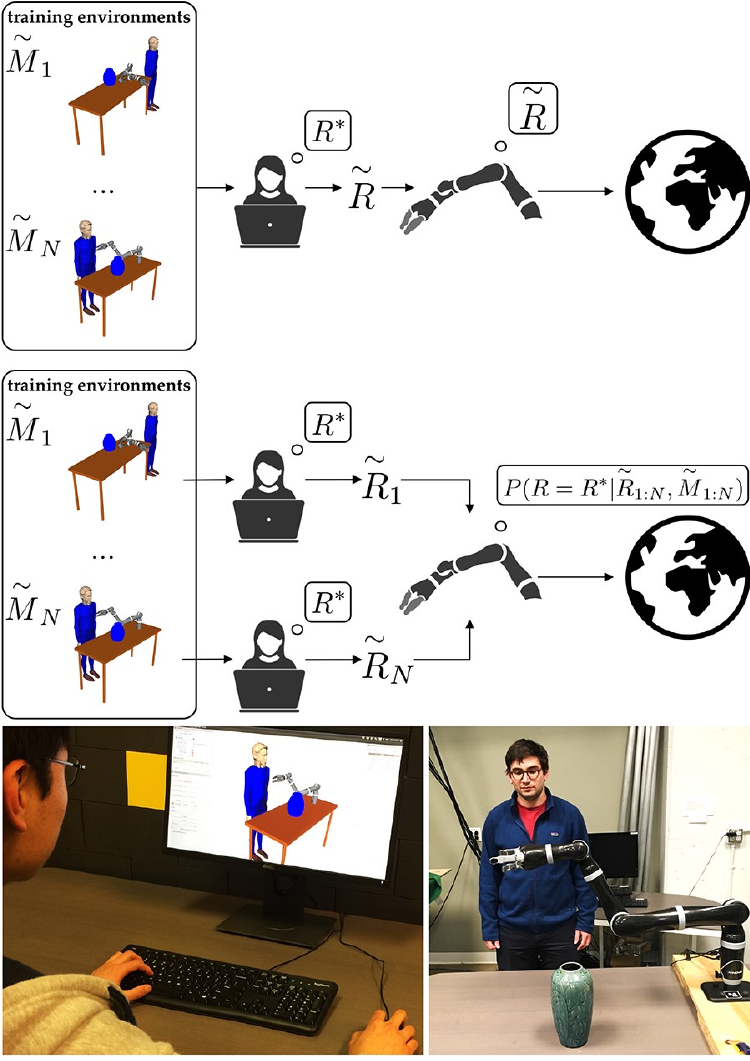}
    \caption{A comparison between the joint reward design and our novel independent reward design approach. \textbf{Top}: Joint reward design involves a single step in which the designer specifies a reward for all environments; the robot directly optimizes this reward in novel environments. \textbf{Middle}: Independent reward design involves designing a separate reward for each environment. We then must combine these into a common reward for the robot to optimize. \textbf{Bottom}: We test independent reward design on motion planning task for a 7-DOF manipulator.}
	\label{fig:frontfig}
\end{figure}

When designing the reward, the goal is to choose a function that guides the robot to accomplish the task in any potential test environment that it might encounter. Typically, the designer considers a representative set of \emph{training} environments, and finds a reward function that induces desirable behavior across all of them, as in \figref{fig:frontfig} (Top). In practice, this can be both challenging and frustrating for the reward designer. The process often results in many iterations of \emph{tuning}, whereby changing the reward function corrects the behavior in one environment, but breaks it in another, and so on.

We posit that designing a good reward function for a single environment at a time is easier than designing one for all training environments in consideration simultaneously. Imagine the task of motion planning in the home. The reward function provided to the planner must correctly encode the desired trade-offs: the robot must stay away from static objects, it should give wider berth to fragile objects (as in \figref{fig:frontfig} (Bottom)), and it needs to keep a comfortable distance from the person, prioritizing more sensitive areas, such as the head \cite{Mainprice2011}. It is easier to find a good trade-off among these desiderata for a single environment, rather than across many environments with different configurations of the robot, human, and surrounding obstacles, simultaneously. 

Unfortunately, it is not immediately clear how to combine \emph{different} reward functions designed on each environment, into a single, common reward function for the task that generalizes to novel environments. \emph{Our observation is that we can treat each independently designed reward function as evidence about the true reward.}%, and infer a posterior distribution over this true reward. %To accomplish this, we employ the idea introduced in~\cite{HadfieldMenell2017} that each independently designed reward function can be taken as an \emph{observation} of the true reward function to infer a posterior over this true reward. 

%\begin{quote}   
% \emph{Our insight is that we can infer a common reward function by treating the independently designed, environment-specific rewards as \emph{observations} about the true reward.}
%\end{quote}

Our contribution is two-fold:

\noindent\textbf{\emph{Independent} Reward Design.} We introduce a divide-and-conquer method for reward design, requiring less effort from the engineer to design a good reward function across multiple environments. The designer specifies a different reward for each environment, as in \figref{fig:frontfig} (Middle). This relaxes the requirement of specifying a single reward function that leads to task completion on all environments. To combine the reward functions, we make a simple assumption: that the behaviors that each reward induces in the respective environment has high reward with respect to the true, common reward function. We construct an observation model from this assumption building on our previous work in inverse reward design \cite{HadfieldMenell2017}, and use Bayesian inference to generate a posterior distribution over the true reward from independently designed rewards.

\noindent\textbf{Analysis with Users. }
We evaluate our independent reward design method in a series of user studies where we ask people to design reward functions in two domains: first, for a grid world navigation problem, as in \figref{fig:gridworldstudy}; and second, for a 7-DOF robot motion planning problem, in an environment containing both a human, as well as fragile and non-fragile objects. Across both domains, we find that independent reward design is significantly faster, with a 51.4\% decrease in time taken on average, when compared to joint design. Users also found the independent process to be easier, with an average 84.6\% increase on the subjective ease of use scale, and the solutions produced were higher quality, incurring an average of 69.8\% less regret on held-out test environments when compared to the baseline joint reward process. 
 
In addition to demonstrating the effectiveness of our independent reward design approach in the primary studies, we formulate a set of sensitivity analysis studies to understand more thoroughly \emph{when} and \emph{why} independent reward design works best. We vary several factors, such as training set size, difficulty level of the training set, as well as fraction of features present in each training environment. We find that independent design performs consistently better, regardless of difficulty level or number of environments: independent design takes 25.7\% less time, and planning with the resulting reward distribution incurs 43.6\% less regret on average. We find that independent design exhibits the strongest advantage when each environment contains only a subset of the possible features relevant to the task, and otherwise performs on par with joint design.

Overall, we are excited to contribute a framework that has the potential to make designing robot reward functions faster, easier, and more reliable in important real-world robotics problems.

\section{Related Work}

%\adnote{took a pass at this:}

\prg{Inverse Reward Design} Our approach draws on
``Inverse Reward Design (IRD)" \cite{HadfieldMenell2017}, a recent method designed to account for cases where the designer specified a reward function that works in training environments (i.e. jointly designed a reward function), but that might not generalize well to novel environments. For example, if the reward function for a self-driving car were designed under clear weather conditions, then the designer may have failed to specify that the car should slow down when the road is wet. IRD interprets the specified reward as an \emph{observation}, constructing a probability distribution over what the true reward might be by looking at the context (the environments) in which the original reward was specified. When combined with risk-averse planning, this posterior over reward functions enables the robot to act conservatively in novel environments, i.e. slow down in rainy conditions.

In this work we have a different goal: we seek to make it easier to specify the reward. Treating the reward functions designed independently across multiple environments as observations enables us to combine them into a generalizable reward function.
% In this work, we use the idea that designed reward functions should be treated as observations, but with a different goal: we seek to make it easier to specify the reward in the first place (via independent reward design). Treating the independently designed reward functions over multiple environments as observations enables us to combine them. Thus, our primary contribution is about designing \emph{better} rewards \emph{faster} by using the insight that we can treat independent rewards as observations of the true reward to combine them. 

%Although our results do show that IRD improves
%performance on novel environments when the reward was designed jointly
%across test environments, independent reward design outperforms this
%IRD-augmented baseline. Thus, our primary contribution is that
%treating a designed reward function as an observation, as in IRD,
%allow us to transition from joint to independent reward design and
%design better rewards \emph{faster}.

\prg{Designing Reward Functions}
The reward function is a crucial component of many planning and learning approaches. In ``Where do rewards come from?"~\cite{singh2010where}, Singh et al. consider the problem of designing a reward function for an autonomous agent. They argue that, if the agent is suboptimal, then it may improve performance to give it a different reward function than the true reward function. In this work, we consider the setting where the \emph{reward designer}, and not the agent, is the suboptimal one. We show that a divide-and-conquer approach to reward design reduces the burden on designers and enables better overall performance.

\cite{shaikh2017design} proposes and evaluates several interfaces for reward specification, eliciting three-way trade-offs from users.  This is complementary to our approach, and could be adapted to collect data for our method. Furthermore, the divide-and-conquer strategy suggests a natural way to scale this approach to collect arbitrary $n$-way trade-offs because we can combine observations from several three-way interactions. 

\cite{MacGlashan2015} presents a system to ground natural language commands to reward functions that capture a desired task. Using natural language as an interface for specifying rewards is complementary to our approach as well. 

\prg{Inferring Reward Functions} 
There is a large body of research on inferring reward functions. An especially active area of study is ``Inverse Reinforcement Learning (IRL)" \cite{Ng2000, Ziebart2008, Syed2007, Evans2016}, the problem
of extracting a reward function from demonstrated behavior. While we share the common goal of inferring an unobserved reward function, we focus on specifying \emph{reward functions}, and not demonstrating \emph{behaviors}, as the means to get there.

Sometimes expert demonstrations do not communicate objectives most efficiently. In complex situations a supervisor may be able to recognize the correct behavior, but not demonstrate it. Sugiyama et al. \cite{Sugiyama2012} use pairwise preferences to learn a reward function for dialog systems. Loftin et al. \cite{loftin2014strategy} perform policy learning when the supervisor rewards \emph{improvements} in the learned policy. Jain et al. \cite{jain2015learning} assume that users can \emph{demonstrate} local improvements for the policy. Christiano et al. \cite{christiano2017deep} generate comparison queries from an agent's experience and ask users for binary comparisons. Sadigh et al. \cite{dorsa2017active} show how to do active information gathering for a similar problem. Daniel et al. \cite{Daniel2015} actively query a human for ratings to learn a reward for robotic tasks. In our view, these are all complementary approaches to elicit preference information from people, and our approach to reward design is another option in the toolkit. 

\section{Independent Reward Design}
\prg{Problem Statement}
The robot must perform a task defined via some \emph{true reward} function, which it does not have access to. The reward designer knows the task and thus knows this true reward function \emph{implicitly}. Unfortunately, reward functions are not easy to explicate \cite{Amodei2016} -- the designer has to turn intuition into a numerical description of the task.

We consider the reward $R : \Xi \rightarrow \mathbb{R}$ to be a function parameterized by the $k$-dimensional vector $\theta \in \mathbb{R}^k$, where $\Xi$ is the set of trajectories, or sequences of states and actions. The true reward function is parameterized by $\thetatrue \in \mathbb{R}^k$. In practice, we may wish to choose a class of functions whose parameters are easily specified by the designer (e.g. linear in features of the states and actions), and for which there exist efficient optimization algorithms.

Both the robot and the designer have access to a set of $N$ training  environments $\{\trainenv_{1:N}\}$. The designer specifies $N$ \emph{proxy} reward functions $\thetaprox_{1:N}$, one for each environment, where we assume that $\thetaprox_{i}$ leads to successful task execution in environment $\{\trainenv_{i}\}$. Specifically, success corresponds to finding the trajectory $\xi \in \Xi$ that achieves maximal \emph{true} reward $R(\xi; \thetatrue)$, which we denote $\xi_{\thetatrue}^*$.

We seek to estimate the true reward parameters $\thetatrue$ given a set of proxy rewards $\thetaprox_{1:N}$ and corresponding training environments $\trainenv_{1:N}$ in which they were specified. 

%\subsection{Approach} 
\label{sec:approach}
\noindent\textbf{Obtaining a Posterior Over Rewards.}
We compute a posterior over rewards given the observed proxies
\begin{multline} \label{eq:posterior}
    P(\theta = \thetatrue | \thetaprox_{1:N}, \trainenv_{1:N}) \propto \\ P(\thetaprox_1 | \theta, \thetaprox_{2:N}, \trainenv_1) \cdots  P(\thetaprox_N | \theta, \trainenv_N)P(\theta). 
\end{multline}
A reasonable simplifying assumption to make, however, is that the proxy rewards $\thetaprox_i$ are conditionally independent, given the true reward $\theta$: the reward designer generates the $i$th proxy reward $\thetaprox_i$ based only on the true reward $\thetatrue$ and the environment $\trainenv_i$, independent of all other proxy rewards and environments. We can therefore factor \eref{eq:posterior} as
\begin{equation} \label{eq:irdposterior}
    P(\theta = \thetatrue | \thetaprox_{1:N}, \trainenv_{1:N}) \propto \left( \prod_{i=1}^N P(\thetaprox_i | \theta, \trainenv_i)
\right) P(\theta). 
\end{equation}

\noindent\textbf{Observation Model.}
We adapt the observation model from Inverse Reward Design (IRD) \cite{HadfieldMenell2017}. We assume that for a given training environment \trainenv, the reward designer wants to choose a proxy $\thetaprox$ that maximizes the \textit{true reward} obtained by a robot following a trajectory that optimizes $\thetaprox$, or $\thetaprox = \argmax_{\theta} R(\xi^*_\theta; \thetatrue)$ subject to the dynamics of $\trainenv$, where $\xi^*_\theta$ is an optimal trajectory in $\trainenv$ under the reward function parameterized by $\theta$. As discussed, however, the reward designer is imperfect; therefore, we model the designer as being only \emph{approximately} optimal at choosing the proxy reward, or
\begin{equation} \label{eq:obs}
	P(\thetaprox | \thetatrue, \trainenv) \propto \exp\left( \beta R(\xi^*_{\thetaprox}; \thetatrue) \right) 
\end{equation}
$\beta$ controls how optimal we assume the reward designer to be.%, and $\xi^*_{\thetaprox}$ is again an optimal trajectory in $\trainenv$ under the proxy reward function parameterized by $\thetaprox$.

\noindent\textbf{Approximation.}
Ultimately, we wish to estimate the posterior \eref{eq:irdposterior}, which requires the normalized probability $P(\thetaprox | \theta, \trainenv)$
\begin{equation}
		P(\thetaprox | \theta, \trainenv) = \frac{\exp\left( \beta R(\xi^*_{\thetaprox}; \theta) \right)}{Z(\theta)}
\end{equation}
where
\begin{eqnarray}
	Z(\theta) & = & \int_{\thetaprox} \exp\left( \beta R(\xi^*_{\thetaprox}; \theta) \right)\,d\thetaprox 
\end{eqnarray}
Unless the space of proxy rewards $\thetaprox$ is small and finite, solving for $Z$ is intractable. Furthermore, computing the integrand requires solving a planning problem to find $\xi^*_{\thetaprox}$, which adds an additional layer of difficulty (this is known as a \emph{doubly-intractable distribution} \cite{Murray2006}). Because of this, we approximate $Z$ using Monte Carlo integration,
\begin{equation}
	\hat{Z}(\theta) = \frac{1}{N}\sum_{i=1}^N \exp \left( \beta R(\xi^*_{\theta_i}; \theta) \right)
\end{equation}
sampled reward parameters $\theta_i$. This assumes that $\thetaprox$ lies in the unit hypercube. In practice, we compute an optimal trajectory $\xi^*_{\thetaprox_i}$ under the proxy reward $\thetaprox_i$ using a domain-specific motion planning algorithm (e.g. discrete or continuous trajectory optimization). %% TODO (Ellis): should talk about how in practice trajectory optimizers typically return a *locally* optimal solution, and this is something we plan to investigate more formally in future work

Finally, we employ the Metropolis algorithm to generate a set of samples from the desired posterior over the true reward in \eref{eq:irdposterior}. 

\noindent\textbf{Using the Posterior for Planning.}
If independent reward design were not available, the designer would simply give the robot a proxy that works across all training environments. Our approach infers a distribution over rewards, as opposed to a single reward that the robot can optimize in new environments. 

In order to use this posterior in planning, we maximize expected reward under the distribution over rewards given by \eref{eq:irdposterior},
\begin{equation}
	\xi^* = \argmax_{\xi} \mathbb{E}_{\theta \sim P(\theta | \thetaprox_{1:N}, \trainenv_{1:N})}[R(\xi; \theta)]
\end{equation}
which is equivalent to planning with the mean of the distribution over reward functions. While this method has the benefit of simplicity, it disregards reward uncertainty while planning. As discussed in \cite{HadfieldMenell2017}, there are other useful ways of using the posterior, such as risk-averse planning. 

\section{The Joint Reward Design Baseline} \label{sec:jointrd}
\prg{Nominal}
Typically, the designer will choose a single reward that induces the desired behavior jointly across all training environments $\trainenv_i$. The robot then optimizes this in novel environments. The implicit assumption here is that the chosen reward matches the designer's true reward \thetatrue.

\prg{Augmented}
Because the reward designer might not be perfect at communicating the true reward to the robot, the nominal jointly designed reward may not be correct. Perhaps more importantly, there are cases when there are multiple reward functions that induce optimal behavior with respect to the true reward on the training environments, yet only a subset of those rewards perform well in the test environments. The comparison between nominal joint design and independent reward design is the practically relevant one, but scientifically it leaves us with a confound: independent reward design combines the idea of independent design with the idea of treating specified rewards as observations. We thus use the latter to augment joint reward design as well: we treat the single reward $\thetaprox$ designed across all training environments $\trainenv_i$ as an observation of the true reward \thetatrue
\begin{equation}
	P(\theta = \thetatrue | \thetaprox, \trainenv_{1:N}) \propto P(\thetaprox | \theta, \trainenv_{1:N}) P(\theta).
\end{equation}
This is very similar to \eref{eq:irdposterior}. The main difference is that the observation model $P(\thetaprox | \theta, \trainenv_{1:N})$ here is conditioned on the entire set of training environments, rather than a single specific environment. This formulation leaves us with a choice of how to define this observation model. 
We use a simple observation model that assumes the designer chooses the same proxy reward $\thetaprox$ in \emph{all} of the environments

\begin{equation}
    P(\thetaprox | \theta, \trainenv_{1:N}) \propto \prod_{i=1}^N P(\thetaprox | \theta, \trainenv_i).
\end{equation}

We approximate this distribution similarly to the independent reward design observation model.

%%%%%%%%%%%%%%%%%%%%%%%
%%%%% GRID WORLDS %%%%%
%%%%%%%%%%%%%%%%%%%%%%%
\section{Evaluating Independent Reward Design \\in Abstract Domains} \label{sec:gridworldstudy}

\subsection{Experiment Design}

\prg{Type of Design Problem}
We begin by evaluating independent reward design in grid worlds, on a set of environments that capture some of the challenges of reward tuning that designers often face in many real robotics problems. 

We start with this domain because it affords us greater flexibility in manipulating the environments and the difficulty of the task, which we will leverage for our sensitivity analysis in \sref{sec:sensitivity}. In the next section, we will test the generalization of the results from this domain to a real robot task.

\begin{figure*}
\centering
    \begin{tabular}{cc}
        \includegraphics[height=0.3\textwidth]{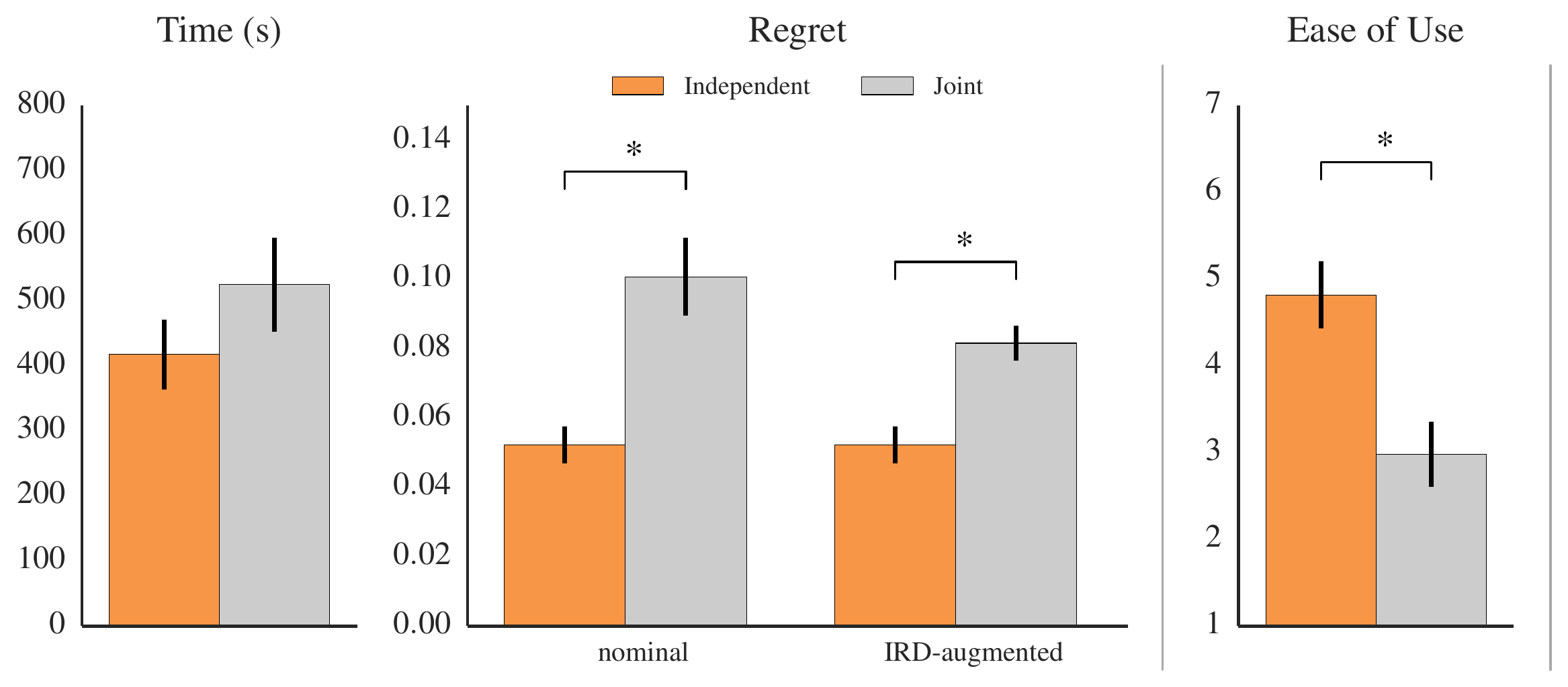}
         & \includegraphics[height=0.3\textwidth]{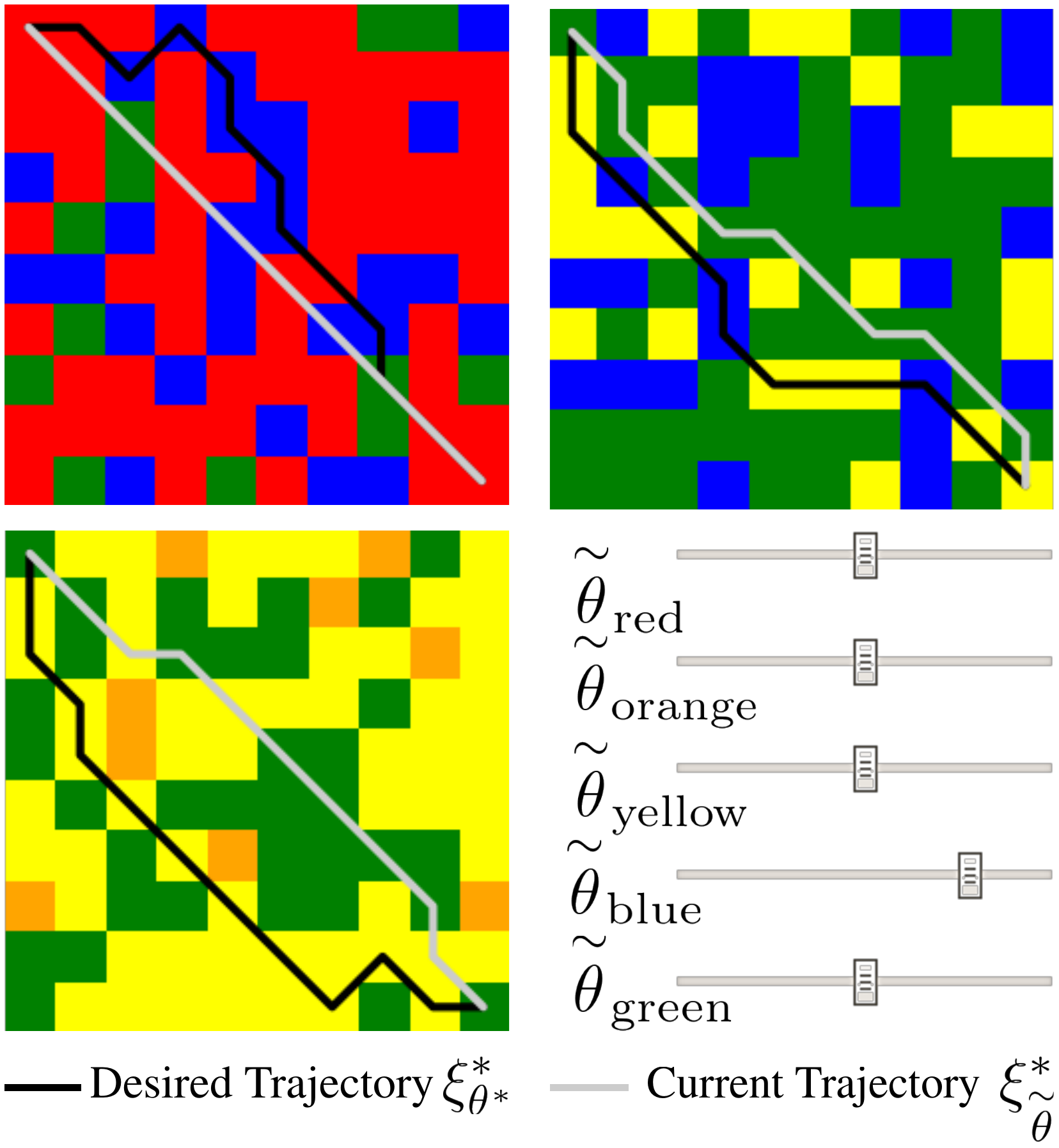}
    \end{tabular}
    \caption{\textbf{Left: }The results of our study in grid worlds indicate that independent reward design is faster, leads to better performance, and is rated as easier to use. \textbf{Right: } Some of the grid world environments used in the study. Participants tuned the parameters of the reward function by using a set of sliders.}
    \label{fig:gridworldstudy}
\end{figure*}

\prg{Problem Domain}
In this experiment, we ask participants to design a reward function for a standard 8-connected grid world navigation task.

For this study, we heuristically chose a set of environments that exhibit several characteristics of a typical, real-world robotics task. 
When designing a reward function, a roboticist typically attempts to sample environments that capture a diverse set of conditions under which the robot operates. For example, when designing a reward for a self-driving car, we choose training environments that exhibit a variety of weather, traffic conditions, road geometries, etc. Similarly, no two of environments in our training set have the same distribution of features/colors. 
An additional characteristic of real-world robotics domains is that not all features will be present in a single environment. For example, we would rarely expect a self-driving car to experience all weather and traffic conditions, road geometries, etc. in a single training environment. Likewise, we chose the set such that not all the features are present in a single environment.  

\prg{Independent Variables} 
We manipulated the reward design process: either independent, or joint. In both cases, we presented the user with the same set of 5 environments, one of which is shown in \figref{fig:gridworldstudy} (Right). For independent reward design, we presented the user with each environment separately. For joint reward design, we presented the user with all environments at the same time.  

In order to evaluate the performance of the algorithms, we needed access to the ground truth reward. Therefore, rather than asking users to design according to some internal preference, we showed them the trajectory that results from optimizing this ground truth reward. We then asked them to create a reward function that incentivizes that behavior.

\prg{Dependent Measures}
\noindent\textbf{\emph{Objective Measures.}}
We measured the time taken by the user to complete the task for each condition. We also measured the quality of the solution produced by reward design using regret computed on a set of 100 randomly-generated environments different from the training set.

Specifically, we employed 2 measures of performance using the output from reward design. The first (regret nominal) measures the regret when planning with the proxy for joint, and with the mean of the posterior over the true reward function for independent, as described in \sref{sec:approach}.

To provide a fair comparison between the two conditions, we applied the machinery developed for generalizing designed rewards to joint reward design. This approach, which we call augmented joint reward design (see \sref{sec:jointrd}), produces a distribution over the true reward function, which allows for more direct comparisons to independent reward design.

This leads to a second measure (regret IRD-augmented), whereby we compute the regret incurred when optimizing the mean of the distribution produced by augmented joint reward design.

\noindent\textbf{\emph{Subjective Measures.}} We used the Likert scale questions in the \hyperref[tab:likert]{Likert Questions} table to design a scale that captures the speed and ease of use for a reward design process. 
 
\prg{Hypothesis}
\noindent\textbf{H1.} \emph{The independent design process will lead to lower regret within a smaller amount of time, and with higher subjective evaluation of speed and ease of use.}

\prg{Participants and Allocation} We chose a within-subjects design to improve reliability, and controlled for the learning effect by putting independent always first (this may have slightly disadvantaged the independent condition). There were a total of 30 participants. All were from the United States and recruited through Amazon's Mechanical Turk (AMT) platform, with a minimum approval rating of 95\% on AMT.  

\subsection{Analysis}
\noindent\textbf{Objective Measures.}
We first ran a repeated-measures ANOVA with reward design process as a factor and user ID as a random effect on the total time and the regret. 

We found that independent outperformed joint across the board: the reward designed using the independent process and the posterior from the IRD-based inference method in \sref{sec:approach} had significantly lower regret ($F(1,30)=10.32$, $p<.01$), and took significantly lower user time to design ($F(1,30)=4.61$, $p=.04$).

Because IRD-based inference is also meant to generalize the designed rewards, we also tested the IRD-augmented joint reward. IRD did improve the regret of the jointly designed reward, as expected. The regret was still significantly lower for independent ($F(1,30)=15.33$, $p<.001$), however.

\figref{fig:gridworldstudy} (Left) plots the results. Supporting our hypothesis, we see that at least for this kind of problem the independent approach enables users to design \emph{better} rewards \emph{faster}.

{\renewcommand{\arraystretch}{1.01}% spaces out entries of table
\begin{table}
  \label{tab:likert}
  \begin{tabular}{l}
    \multicolumn{1}{c}{\textbf{Likert Questions}}  \\
    \hline
    \multicolumn{1}{p{8cm}}{Q1: It was easy to complete [process].} \\
    \multicolumn{1}{p{8cm}}{Q2: I had a harder time with [process].} \\
    \multicolumn{1}{p{8cm}}{Q3: [process] was fast to complete.} \\
    \multicolumn{1}{p{8cm}}{Q4: [process] took fewer runs.} \\
    \multicolumn{1}{p{8cm}}{Q5: [process] was frustrating.} \\
    \multicolumn{1}{p{8cm}}{Q6: I went back and forth between different slider values a lot on [process].} \\
    \hline
  \end{tabular}
\end{table}}

\noindent\textbf{Subjective Measures.} The inter-item reliability of our scale was high (Cronbach's $\alpha=.97$). We thus averaged our items into one rating and ran a repeated measures ANOVA on that score. We found that independent led to significantly higher ratings than joint ($F(1,30)=33.63$, $p<.0001$). The mean rating went from $2.35$ for joint to $5.49$ for independent.

%\begin{quote}
\prg{Summary} \emph{Overall, not only do users perform better with the independent process and faster, but they also subjectively prefer it over the joint process.}   
%\end{quote}

%%%%%%%%%%%%%%%%%%%%%%
%%%%% JACO STUDY %%%%%
%%%%%%%%%%%%%%%%%%%%%%
\section{Evaluating Independent Reward Design\\ for a Robot Manipulator} \label{sec:jacostudy}

\subsection{Experiment Design}

\begin{figure*}
\centering
    \begin{tabular}{ccc}
    \includegraphics[height=0.23\textwidth]{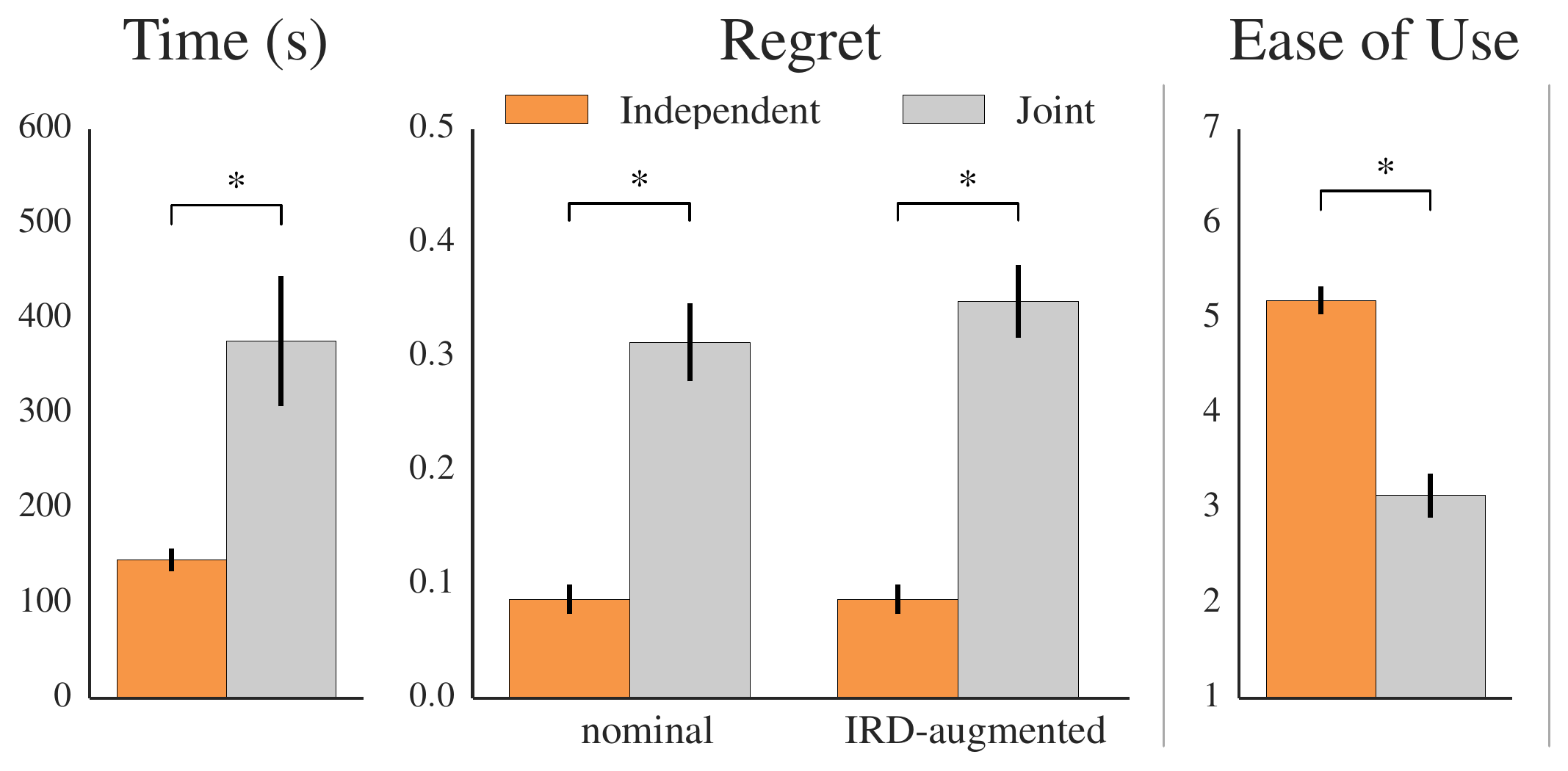} & 
    \includegraphics[height=0.23\textwidth]{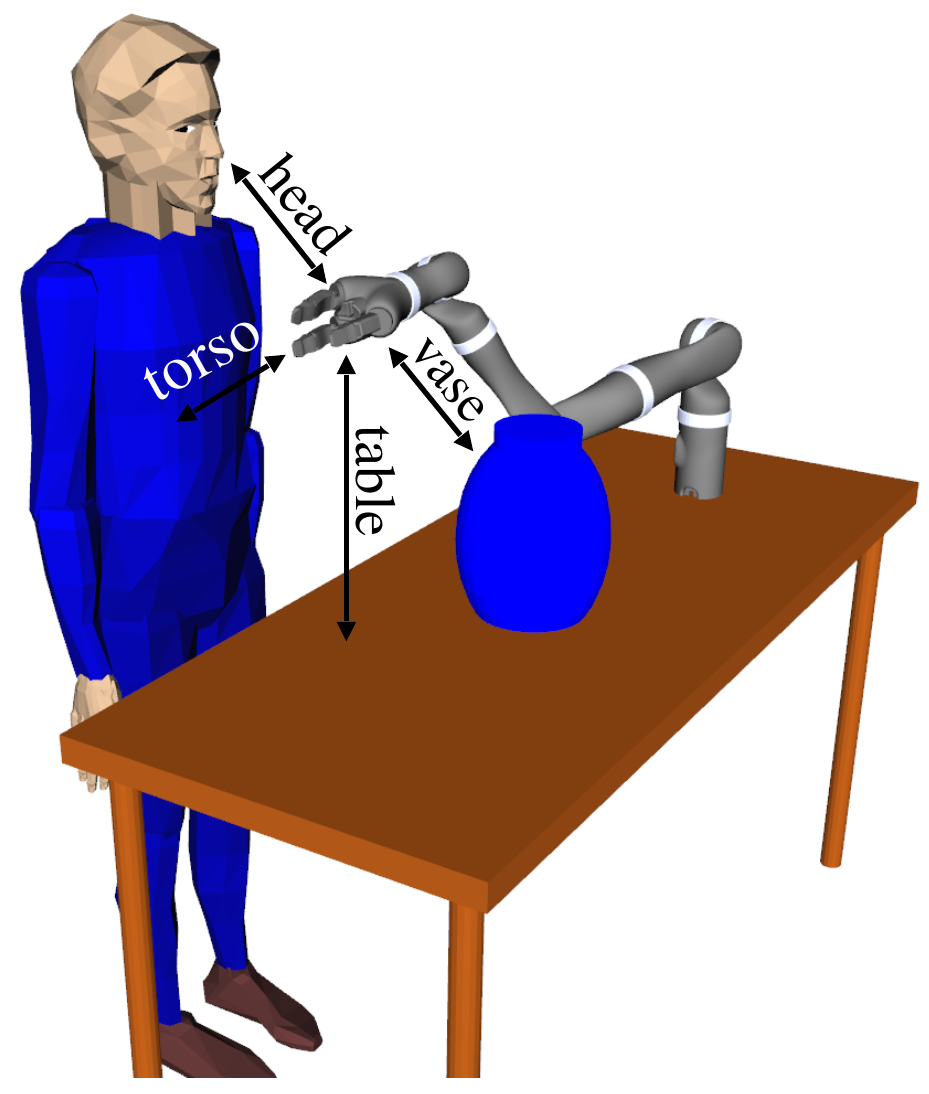} & 
    \includegraphics[height=0.23\textwidth]{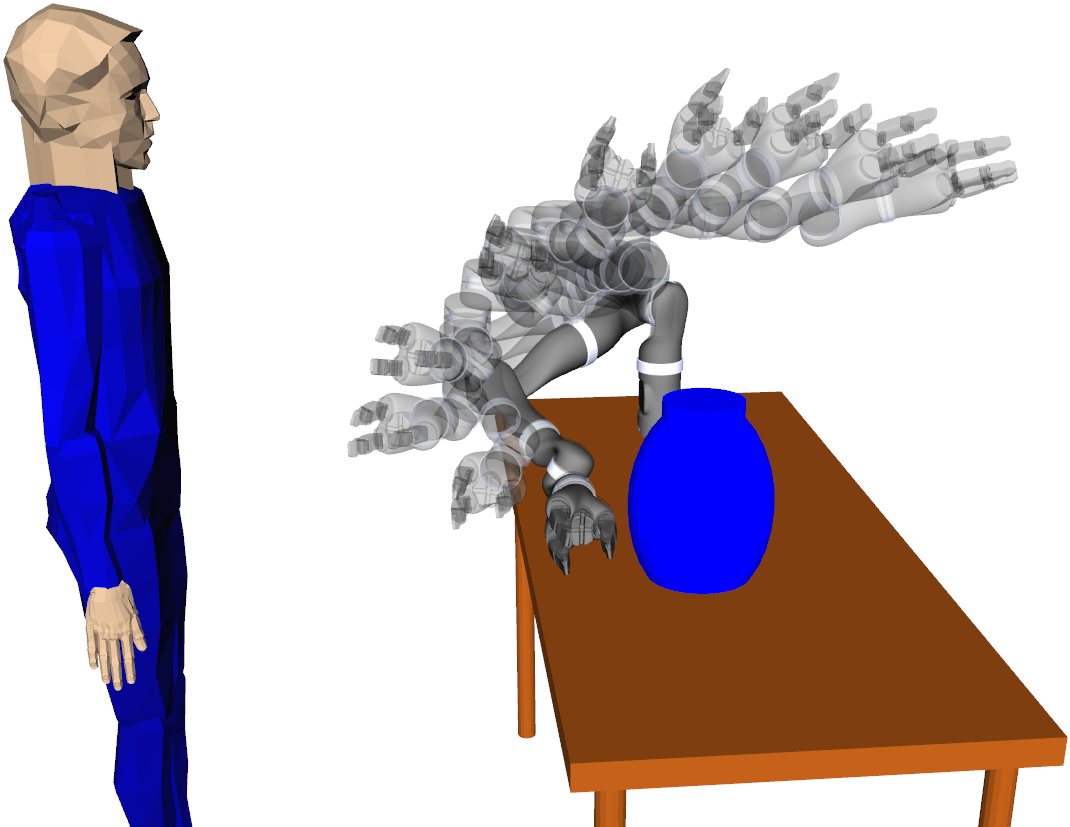}
    \end{tabular}
    \caption{\textbf{Left: }The results of our study on the Jaco 7-DOF arm are consistent with the previous study in grid worlds, further illustrating the potential benefits of independent reward design. \textbf{Right: } Two of the environments used in this study. Participants were asked to tune the weights on 4 features to achieve a desired trajectory: RBF distance to the human's head, torso, RBF distance to the vase, and distance from the table. After changing changing these weights, users were able to forward-simulate the trajectory to evaluate the induced behavior, as shown on the right.} 
    \label{fig:jacostudy}
\end{figure*}

\prg{Problem Domain}
Following the promising results of our grid world experiments, we sought to apply independent reward design to a robot motion planning problem. In this experiment, we examine reward design for the Jaco 7-DOF arm planning trajectories in a household environment. We employ TrajOpt \cite{Schulman2014}, an optimization-based motion planner, within a lightweight ROS/RViz visualization and simulation framework. Each environment contains a different configuration of a table, a vase, and a human. We used linear reward functions. Our features were radial basis function (RBF) distances from the end-effector to: the human's torso, the human's head, and the vase. We had an additional feature indicating the distance from the end-effector to the table. These features are illustrated in \figref{fig:jacostudy}.

Participants specified the weights on the 4 features by tuning sliders, as in the grid worlds study in \sref{sec:gridworldstudy}. We asked participants to design a reward that induces a trajectory close to a ground truth trajectory that is visualized. This ground truth trajectory is the result of optimizing a ground truth reward function that is hidden from the participants. 

\prg{Variables}
We used the same independent and dependent variables as in the grid world study in \sref{sec:gridworldstudy}.

% \prg{Independent Variables} 
% We manipulated the same variables as in \sref{sec:gridworldstudy}. We presented the user with 3 training environments, similar to the example shown in \figref{fig:jacostudy} (Right).

% \prg{Dependent Measures}

% \noindent\textbf{\emph{Objective Measures.}}
% We employed the same objective measures as in \sref{sec:gridworldstudy}, with the exception of a measure of regret under risk-averse planning.

% \noindent\textbf{\emph{Subjective Measures.}}
% We employed the same subjective measures as in \sref{sec:gridworldstudy}.
 
\prg{Hypothesis}
\noindent\textbf{H2.} \emph{The independent design process will lead to lower regret within a smaller amount of time, and with higher subjective evaluation of speed and ease of use.}

\prg{Participants and Allocation} We chose a within-subjects design to improve reliability, and counterbalancing to mitigate strong learning effects observed in the pilot. There were a total of 60 participants. All were from the United States and recruited through AMT with a minimum approval rating of 95\%.  

\subsection{Analysis}

\noindent\textbf{Objective Measures.} 
Our results in the manipulation domain are analogous to the grid world domain. We found a significant decrease in time ($F(1,59)=10.97$, $p<.01$) and regret (both nominal, $F(1,59)=38.66$, $p<.0001$, and augmented, $F(1,59)=54.68$, $p<.0001$) when users employed independent for designing rewards, as illustrated in \figref{fig:jacostudy}.

\noindent\textbf{Subjective Measures.} 
We found that users found independent to be significantly easier to use than joint in designing rewards ($F(1, 59) = 40.76$, $p<.0001$), as shown in \figref{fig:jacostudy}.

%\begin{quote}
\prg{Summary} \emph{Overall, our results for manipulation follow those in the grid world domain: not only do users perform better with the independent process and faster, but they also subjectively prefer it over the joint process.}   
%\end{quote}

\section{Sensitivity Analysis}
\label{sec:sensitivity}

From a practical standpoint, our experiments thus far showcase the advantages of independent reward design over the traditional approach of designing rewards jointly. We saw equivalent results in the grid world and the motion planning task. From a scientific standpoint, however, there are still open questions: we need push the limits of this method and investigate exactly what factors of a design problem lead to its advantages. We conduct three additional experiments. We chose to conduct this sensitivity analysis in the abstract grid world domain of \sref{sec:gridworldstudy} to make it easier to manipulate properties of the environments that we present to participants.

\begin{figure*}
\centering
    \centering
    \includegraphics[width=\textwidth]{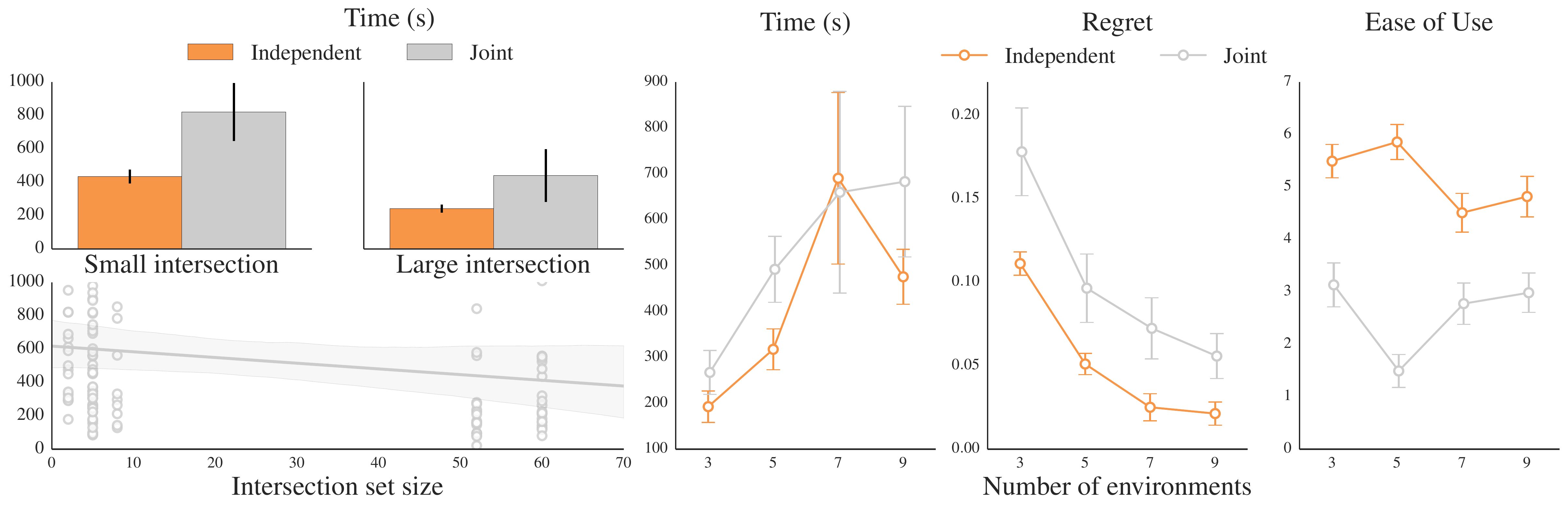}
    \caption{\textbf{Left: }Across small or large feasible sets and intersections sizes, independent still outperforms joint design. A smaller intersection size makes the problem harder, as expected. \textbf{Right: } As the number of environments increases, we still see benefits in regret and ease of use, but the time difference is less pronounced. }
    \label{fig:factorial}
\end{figure*}
% ER: Dylan pointed out that there is an outlier in the time data when # envs = 7; we need to mention this in the paper to address the otherwise unexplained spike at that point

%%%%%%%%%%%%%%%%%%%%%%%%%%%%%
%%%%% FEASIBLE SET SIZE %%%%%
%%%%%%%%%%%%%%%%%%%%%%%%%%%%%
\subsection{Feasible Rewards Set Size}

In this experiment, we characterize the difficulty of a reward design by examining the \emph{feasible rewards set} $\mathcal{F}_\epsilon(M, \xi)$ for a problem: the set of reward functions that produce the desired trajectory $\xi$ in a given environment, within some tolerance $\epsilon \geq 0$. That is, 
\begin{equation}
\mathcal{F}_\epsilon(M, \xi) = \{ \theta \in \mathbb{R}^k : d(\xi, \xi_\theta^*) \leq \epsilon \}, 
\end{equation}
where $\xi_\theta^* = \text{argmax}_\xi R(\xi; \theta)$ subject to the dynamics of environment $M$, $d : \Xi \times \Xi \rightarrow \mathbb{R}$ is a distance metric on $\Xi$, and $\epsilon \geq 0$ is a small constant. We vary both the size of $\mathcal{F}_\epsilon(M_i, \xi)$, as well as the intersection over all $\mathcal{F}_\epsilon(M_i, \xi)$, for $i = 1, \dots, N$. For environments with discrete state and action spaces, we let $\epsilon = 0$, so that trajectories in the feasible set must match exactly the true trajectory.

%We first vary the size of this set per environment. If this set is small for an environment, then finding the correct rewards is hard. We also vary the intersection size of feasible rewards sets across all training environments. If this intersection is small, then finding a single reward function that elicits the correct behavior across all environments is hard. 

%We've seen in the main study what happens in an illustrative case-- one where it is might be relatively easy to design a reward for each environment (large feasible sets); but where it might be relatively difficult to design a reward for all environments at once (small intersection of feasible sets). We conducted a follow-up study where we purposefully changed these parameters. 

% \subsection{Experiment Design}
\prg{Independent Variables}
We manipulate the reward design process (joint vs. independent), the average feasible set size over the set of environments (large vs. small), and the size of the intersection of the feasible sets for all environments (large vs. small).

We chose the categories of feasible sets using thresholds selected by examining the feasible set sizes and intersection sizes in the previous studies. To choose the environment sets for each of the 4 categories, we followed a simple procedure. We sampled sets of 5 environments, and computed the feasible set size and intersection size for each set. Then we chose 1 set from each category randomly to use in this study.

%% DHM -- this is unclear to me, seems like we aren't giving enough detail about how we bucketed things into big/small

%\subsubsection{Dependent Measures}
%We use the same measures as before.

\prg{Hypotheses}
%We believe that the size of the feasible set of an environment negatively correlates with how complex users will find designing a reward for that environment:

\noindent\textbf{H3.} \emph{The size of the feasible set for an environment negatively affects the amount of time taken for that environment in the independent condition (i.e. larger feasible set leads to less time spent designing the reward). }

%Similarly, we believe that when designing a joint reward, the size of the intersection of the feasible sets for each environment negatively correlates with how complex it is to design the reward:

\noindent\textbf{H4.} \emph{The size of the intersection of feasible sets negatively affects the amount of time taken in the joint condition (again, larger intersection set means smaller amount of time).}

We still believe that on average across this variety of environments, independent leads to better results:

\noindent\textbf{H5.} \emph{Independent will still lead to lower time and regret while receiving higher subjective ratings.}

\prg{Participants and Allocation} Same as the study in \sref{sec:gridworldstudy}, except now we recruited 20 participants per environment set for a total of 80.

\prg{Analysis}
In this experiment, we attempt to characterize difficulty of reward design on a set of environments by two properties: the feasible reward set size for each individual environment, and the intersection of feasible reward sets across the training set. Our results support the hypotheses that environments with \emph{large} feasible reward sets are typically \emph{easier}, and joint performs best for training sets with \emph{large} intersections of feasible reward sets. Overall, however, independent still performs well in all cases. 

To test \textbf{H3}, we fit the amount of time spent for each environment in the independent condition by the size of the feasible set for each user and environment, and found a significant negative effect as hypothesized ($F(1,568)=14.98$, $p<.0001$). This supports \textbf{H3}. \figref{fig:factorial} (Bottom) shows a scatterplot of time by feasible set size.

To test \textbf{H4}, we fit the amount of time spent designing the joint reward by the size of the intersection of feasible sets. There was a marginal negative effect on time ($F(1,113)=2.55$, $p=.11$). This provides some partial support to \textbf{H4}.

To test \textbf{H5}, we ran a fully factorial repeated-measures ANOVA for time and regret using all 3 factors: design process, average size of feasible set (large vs. small), and size of intersection of feasible sets (large vs. small). 

For time, we found two significant effects: process and intersections size. Independent indeed led to significantly less time taken \emph{even across this wider set of environments} that is meant to balance the scales between independent and joint ($F(1,163)=10.27$, $p<.01$). And indeed, larger intersection size led to less time ($F(1,163)=8.93$, $p<.01$).

For regret, we again found a significant effect for process, with independent leading to significantly lower regret ($F(1,163)=19.66$, $p<.0001$). Smaller feasible sets and smaller intersections led to significantly lower regret too ($F(1,163)=38.69$, $p<.0001$ for feasible sets and $F(1,163)=8.1$, $p<.01$ for intersections). While smaller feasible reward sets are typically more challenging for the reward designer, they are in fact more informative -- often smaller sets mean that when designers identify the desired reward, there are few others that would induce the same desired behavior. Informally, this means that this recovered reward is more likely to be close to the true reward, and hence will generalize well to new environments and induce lower regret. But there were also interaction effects: between process and feasible set size, and between feasible set size and intersection size. We did posthocs with Tukey HSD. We found that while independent improved regret significantly for large feasible sets (the setting where independent works best, because each environment is easy; $p<.0001$), the improvement was smaller and no longer significant for the small feasible sets. The interaction between the two size variables revealed that when both the sets and their intersection are large, the regret is the highest; when they are both small, the regret is lowest; when one of them is small and the other large, the regret lies somewhere in between the other two conditions. 

The results for both time and regret support \textbf{H5}, and the results for time are summarized in \figref{fig:factorial} (Left).

%\begin{quote}
\prg{Summary} \emph{Overall, our result from the main study generalized to this wider range of environments: independent still lead to less time taken and lower regret. }   
%\end{quote}
However, it does seem like environments with very small feasible sets would be difficult with the independent process, almost as difficult as with joint. Independent will work best when each environment is simple, but getting a reward that works across \emph{all} environments is hard. 

%%%%%%%%%%%%%%%%%%%%%%%%%
%%%%% FEATS PER ENV %%%%%
%%%%%%%%%%%%%%%%%%%%%%%%%
\subsection{Number of Features per Environment}

\begin{figure}
    \centering
    \includegraphics[width=\columnwidth]{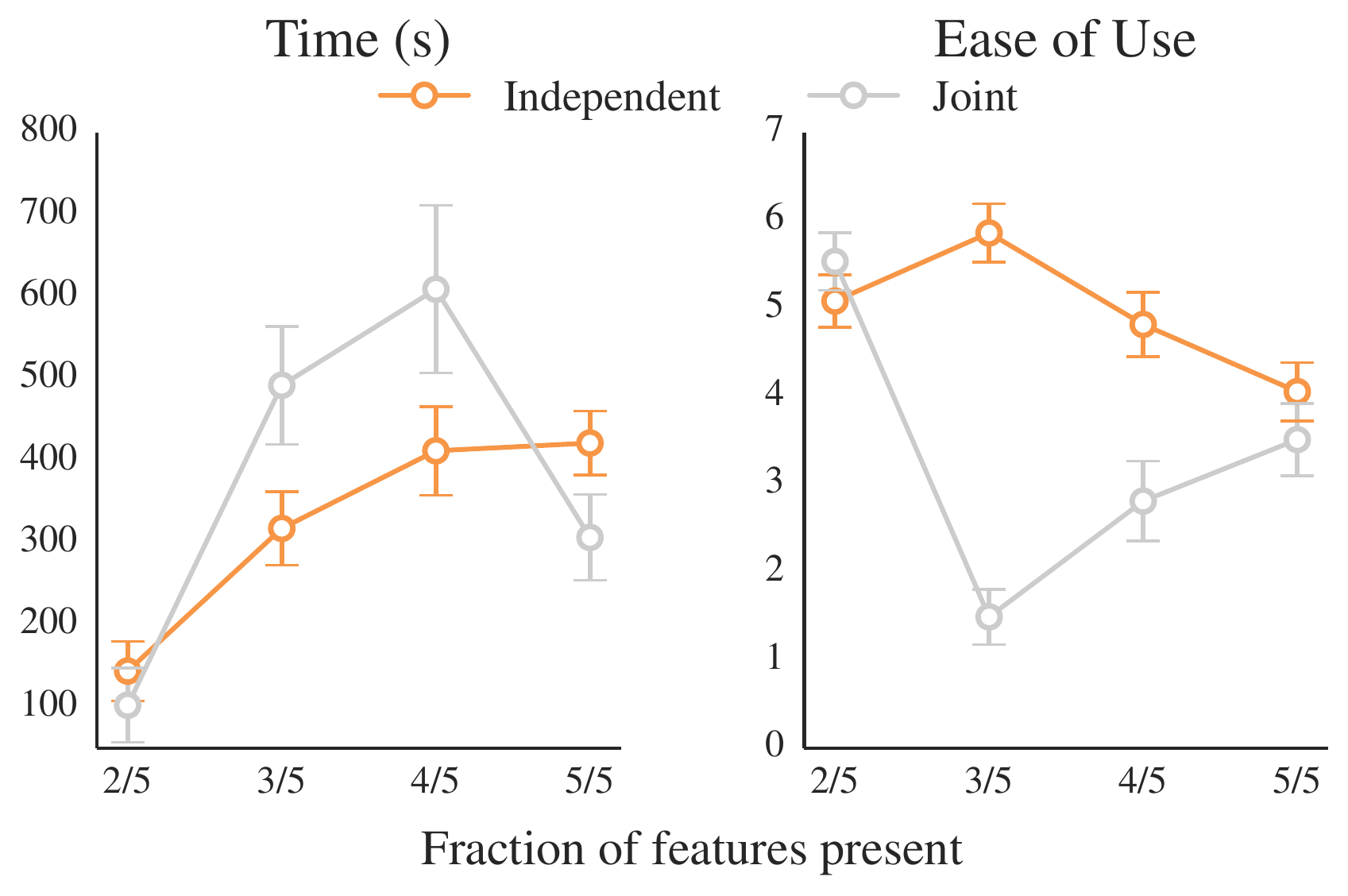}
    \caption{Independent reward design performs best when around half of the total number of relevant features are present in each environment.}
    \label{fig:fracfeats}
\end{figure}

This study investigates the effectiveness of our approach as the number of features present in each environment changes. 

%\subsection{Experiment Design}
\prg{Independent Variables}
We manipulated the fraction of the total features $F$ present in each environment, varying from $F=2/5$ to $F=5/5$, with 5 possible relevant features in this domain (i.e. 5 different terrains in the grid world). %Because we are limited to 5 features, we are actually studying the effect of varying the \emph{fraction} of relevant features per environment on the performance of reward design. 

In total, we had 5 sets of environments (4 on top of our set from \sref{sec:gridworldstudy}). We also manipulated the reward design process as before.

\prg{Hypothesis}
\noindent\textbf{H6.} \emph{Independent reward design outperforms joint when there are some, but not all of the total number of features possible per environment.}
We expect to see that independent has performance on par with joint when all features are present in each environment -- then, the designer has to consider the effect of all features on the induced behavior, and hence must specify weights on \emph{all} features for \emph{all} training environments, the same as in the joint condition. We also expect the same performance between the two conditions when there are only 2 of 5 features present in each environment, because identifying the trade-offs between features becomes trivial.

\prg{Participants and Allocation} Same as the study in \sref{sec:gridworldstudy}, except now we recruited 20 participants per environment set for a total of 80.

\prg{Analysis} Taken as a whole, the data aligns well with our hypothesis. Independent tends to outperform joint with a moderate fraction of the total number of features (3/5 or 4/5), but the two methods perform about the same when the minimum (2/5) or maximum (5/5) fraction of features are present. These results are summarized in \figref{fig:fracfeats}.

We ran a repeated measures factorial ANOVAs, with results largely supporting \textbf{H6}. The effect was most clear in the subjective measures, where we saw an interaction effect between the two factors on our ease of use scale-- the difference between user preference on $F=3/5$ and $F=4/5$ was so strong, that even the Tukey HSD posthoc, which compensates for making 28 comparisons, found a significant improvement of independent over joint for those feature numbers ($p<.0001$ and $p<.02$). For other measures the difference on these two features was not strong enough to survive this level of compensation for multiple comparisons, but of course planned contrasts of independent vs. joint on each of the number of features support the improvement in both regret and time.  

%\begin{quote}
\prg{Summary}\emph{Overall, independent outperforms joint when each environment only contains a subset of the total number of features relevant to the task.}
%\end{quote}

%%%%%%%%%%%%%%%%%%%%%%%
%%%%% NUM OF ENVS %%%%%
%%%%%%%%%%%%%%%%%%%%%%%
\subsection{Number of Environments}
Our final experiment varies the number of environments used in the reward design process.

\prg{Independent Variable} 
We manipulated the number of environments in the training set. Starting from the set of environments used in the primary study (\sref{sec:gridworldstudy}), we either randomly removed or added environments to achieve set sizes of 3, 7, and 9. When adding, we implemented a rejection sampling algorithm for heuristically avoiding environment sets that would be trivially-solved by a reward designer.

\prg{Hypothesis}
We did not have a clear hypothesis prior to the study: while increasing the number of environments should make joint reward design more difficult, the effort required for independent reward design should scale linearly with the number of environments as well.

\prg{Participants and Allocation} Analogous to \sref{sec:gridworldstudy}, except we recruited 20 users per condition, for a total of 80 users.  

\prg{Analysis} We conducted a repeated measures factorial ANOVA with number of environments and process as factors, for each of our dependent measures. We found that the number of environments influenced many of the metrics. As shown in \figref{fig:factorial} (Right), independent consistently outperformed joint in terms of regret ($F(1,161)=9.68$, $p<.01$) and ease of use ($F(1,161)=20.65$, $p<.0001$). We also saw that the time taken went up with the number of environments ($F(3,159)=4.94$, $p<.01$); this is unsurprising, as designing a reward for more environments should require more effort.

Furthermore, we observe that independent is most time efficient with respect to joint when there are a moderate number of environments (in this study, 5). If there are too few environments (in this study, 3) or too many environments (in this study, 7 or 9), then the differences in time taken are not significant. This follows our intuition about in what regimes independent or joint reward design should be used.

%This follows our intuition-- if there are only a small number of environments to consider simultaneously, then joint should be an effective strategy for designing a reward; alternatively, if there are too many, then the advantages of divide-and-conquer diminish as the number of reward design ``subproblems'' to be solved grows.  

%\begin{quote}
\prg{Summary} \emph{Overall, we observe that while the reward recovered by independent induces lower regret and independent is consistently rated as easier to use, there is less of an advantage with respect to reward design time if the number of environments is very small or very large.}

%\emph{Overall, we see a mixed effect for measures of user effort. In the limit of a large number of environments, joint reward design leads to fewer calls to the planner but independent reward design leads to less total time taken.} 

%that as the number of environments that the user must consider increases, there is less of a clear distinction in terms of effort, yet independent still leads to significantly lower regret and higher ease of use.}
%\end{quote}

\section{Discussion}

\noindent\textbf{Summary.}
We introduced an approach to dividing the reward design process into ``subproblems'' by enabling the designer to individually specify a separate reward function for each training environment. The robot combines the individual rewards to form a posterior distribution over the true reward, and uses it to plan behavior in novel environments. 

We showed that our independent reward design approach performs better than traditional reward design: not only did users find independent design easier to use, but it also resulted in a better quality solution. Furthermore, through a series of stability analysis experiments, we found that our method works best when specifying a complex reward function for a set of environments can be broken down into specifying simpler reward functions on each environment. This approach works well in robotics domains where the cycle of planning and simulation/testing on hardware is slow or computationally expensive, and hence fewer iterations of reward tuning is necessary. 

\noindent\textbf{Limitations and Future Work.}
One limitation of our experiments was that we provided users with the desired trajectory, rather than asking for them to come up with it, as would be the case in real-world reward design. We did this to control for differences is subjective interpretations of task completion across users. In future work, we plan explore independent reward design ``in the wild''. 

Further, we see that independent reward design is easiest when users are given a moderately-sized set of training environments. In many real-world robotics settings, there are in fact \emph{very large} sets of training environments. In future work, we are interested in exploring how to choose an appropriate subset from a larger set of environments, so as to maximize the solution quality.

Building on the promising results on a robot manipulation task, we are excited about the potential impact of our approach to designing rewards for a broader class of robotics problems. Designing reward function for use in planning for self-driving cars that must operate safely and comfortably across a variety of environments is challenging. We believe that independent reward design could be a helpful tool for engineers to design such rewards with greater ease.

%Designing a good reward function to plan for self-driving cars in particular suffer from a very similar challenge in tuning reward parameters across environments, and independent reward design could be a helpful tool for engineers to designing these reward functions with greater ease.

\section{Acknowledgements}

This work was partially supported by AFOSR FA9550-17-1-0308 and The Open Philanthropy Foundation.

\bibliographystyle{plainnat}
% \bibliography{bibliography} 

\end{document}